\documentclass[sigconf]{acmart}
\AtBeginDocument{%
  }

\setcopyright{acmlicensed}
\copyrightyear{2026}
\acmYear{2026}
\setcopyright{cc}
\setcctype{by}
\acmConference[MMSports '26]{9th Int. Workshop on Multimedia Content Analysis in Sports}{November 10--14, 2026}{Rio de Janeiro, Brazil}
\acmBooktitle{9th Int. Workshop on Multimedia Content Analysis in Sports (MMSports '26), November 10--14, 2026, Rio de Janeiro, Brazil}
\acmDOI{10.1145/3841455.3841533}
\acmISBN{979-8-4007-2944-7/2026/11}

\usepackage{graphicx}
\usepackage{booktabs}
\usepackage{algpseudocode}
\usepackage[ruled,vlined]{algorithm2e}
\usepackage{booktabs}
\usepackage{multirow}
\usepackage{multicol}
\usepackage{tabularx}
\usepackage{subcaption}
\usepackage{balance}

\acmSubmissionID{9}

\begin{document}

\title{Audio for Sports Highlight Detection: A Comparative Empirical Study}
\author{Hao Xu}
\affiliation{%
  \institution{Deakin University}
  \city{Melbourne}
  \country{Australia}
}
\additionalaffiliation{%
  \institution{Dolby Laboratories Inc.}
  \city{Sydney}
  \country{Australia}
}
\email{august.xu@research.deakin.edu.au}

\author{Meenakshi Sarkar}
\affiliation{%
  \institution{Dolby Laboratories Inc.}
  \city{Bangalore}
  \country{India}
}
\email{meenakshi.sarkar@dolby.com}

\author{Vishnu Raj}
\affiliation{%
  \institution{Dolby Laboratories Inc.}
  \city{Bangalore}
  \country{India}
}
\email{vishnu.raj@dolby.com}

\author{David Gunawan}
\affiliation{%
  \institution{Dolby Laboratories Inc.}
  \city{Sydney}
  \country{Australia}
}
\email{dguna@dolby.com}



\begin{abstract}
Sports highlight detection aims to identify the most exciting and meaningful moments from long sports videos. While existing methods often emphasize visual or visual-language representations, sports videos contain rich audio cues, including commentator speech, crowd reactions, whistles, ball impacts, and referee calls. In this work, we revisit the role of audio in sports highlight detection and ask a simple question: how far can audio alone go? We construct lightweight audio-only baselines using pretrained audio representations and compare them with visual-only and audio-visual methods on the SV-Highlights benchmark. Surprisingly, our audio-only GRU baseline achieves strong performance and outperforms several existing audio-visual methods under our supervised evaluation setting. Furthermore, a simple audio-visual fusion baseline achieves the best performance across all metrics, indicating that audio and visual cues provide complementary information. To better understand the contribution of audio, we conduct source-separated analysis and show that vocal/commentary audio is more informative than background-only audio, while their combination performs best. We also analyze interpretable audio cues and find that highlight clips exhibit higher RMS loudness, peak loudness, and mid-frequency energy than non-highlight clips, although substantial distribution overlap indicates that loudness alone is insufficient. Our findings suggest that audio is an underexplored but highly informative modality for sports highlight detection and should be treated as a primary signal rather than merely an auxiliary cue.
\end{abstract}



\maketitle
\section{Introduction}
\label{sec:intro}
Sport is one of the largest and most influential media industries worldwide, with broadcasting forming a central component of its commercial and cultural value \cite{xu2025deeplearningsportsvideo}. Beyond live broadcasting, sports highlights have become an increasingly important format for content consumption, allowing viewers to quickly access the most exciting and meaningful moments without watching an entire match or event \cite{lee2026svhighlights, diaz2025soccerhigh}, meanwhile enabling many downstream tasks including video summarization, recommendation, editing and browsing \cite{badamdorj2021joint}.

Sports highlight detection is an important problem in video understanding. Unlike generic video classification, highlight detection requires models to identify not only what happens in a video, but also whether an event is sufficiently important, exciting, or consequential to be considered as a highlight \cite{della2025automated}. This requires temporal understanding beyond isolated frames or short clips, since the difference between a routine play and a highlight-worthy moment often depends on the surrounding context, game state, audience reaction, and event progression.

\begin{figure}
    \centering
    \includegraphics[width=1\linewidth]{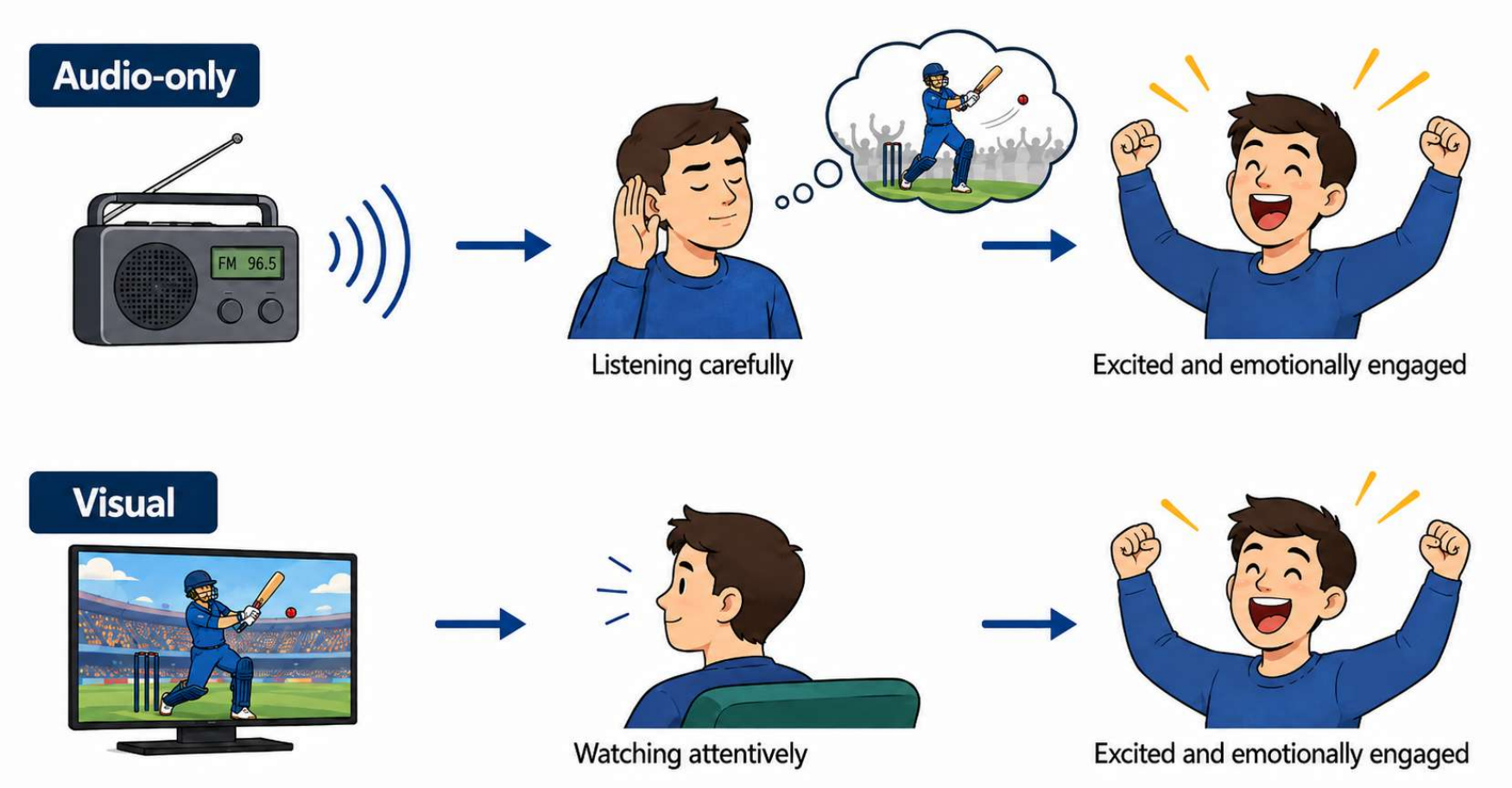}
    \caption{Motivation for studying audio in sports highlight detection.
Both audio-only and visual perception can convey sufficient event
and affective information for audiences to recognize and respond to
important sports moments. This motivates our investigation of how
effectively audio alone can support highlight detection.}
    \label{fig:motivation}
\end{figure}

Compared with general video highlight detection, sports videos contain rich and highly informative audio cues. Sounds such as ball impacts, whistles, crowd cheering, commentator speech, referee calls, and audience reactions often provide direct evidence of important game moments. These cues are intuitively useful for understanding sports events. However, many existing highlight detection methods still rely predominantly on visual representations \cite{badamdorj2021joint, ma2025ms, moon2023query, gordeev2026saliency, lin2023univtg, sun2024tr}, while audio is either ignored, treated as an auxiliary signal, or fused in a relatively simple manner.

This motivates a natural question: how far can audio alone go in sports highlight detection? Interestingly, before the television broadcasting era, sports were widely consumed through radio, where audiences followed and enjoyed games using audio-only descriptions and reactions as shown in Figure \ref{fig:motivation}. This suggests that audio may carry substantially more semantic and affective information than is typically assumed in current video highlight detection systems.

In this paper, we revisit the role of audio in sports highlight detection. Instead of treating audio as a secondary modality, we systematically investigate whether audio-only temporal modelling can identify sports highlights and how it compares with visual and audio-visual approaches. Our study shows that audio-only models using pretrained audio representations and lightweight temporal modelling are surprisingly competitive with existing audio-visual methods. Furthermore, simple audio-visual temporal fusion further improves performance, indicating that audio provides strong highlight-relevant cues that remain underutilized in current approaches.

In summary, this paper makes the following contributions:
\begin{itemize}
    \item We revisit the role of audio in sports highlight detection and show that audio-only temporal modelling is surprisingly competitive with existing audio-visual highlight detection methods.
    \item We introduce simple but strong audio-only and audio-visual temporal baselines using pretrained audio and visual representations.
    \item We conduct source-separated audio analysis to examine the respective contributions of vocal/commentary cues and background sounds, including crowd reactions, whistles, and impacts.
    \item We demonstrate that temporal modelling is critical for audio-based highlight detection, with recurrent temporal models consistently outperforming clip-independent baselines.
\end{itemize}

\section{Related Work}

\textbf{Moment Retrieval and Highlight Detection.}
Highlight detection (HD) and moment retrieval (MR) are increasingly studied together in video understanding. MR aims to localize a temporal moment in a video given a natural-language query, whereas HD focuses on estimating the saliency or importance of each clip within a video~\cite{sun2024tr}. The joint formulation of MR and HD was popularized by QVHighlights~\cite{lei2021detecting}, which introduced Moment-DETR, a DETR-based framework that retrieves query-relevant moments while simultaneously assigning highlight scores to video clips.

Following this formulation, many recent works have adopted DETR-style architectures~\cite{carion2020end} for joint MR\&HD. QD-DETR~\cite{moon2023query} introduces a query-dependent video representation module, making moment localization more explicitly conditioned on the user query. MH-DETR~\cite{xu2024mh} incorporates pooling operations into the encoder and introduces cross-modal interaction modules for joint MR\&HD. TR-DETR~\cite{sun2024tr} argues that previous methods mainly focus on improving multimodal feature discrimination and interaction under a multi-task learning framework, while overlooking the reciprocal relationship between MR and HD. It therefore proposes a task-reciprocal Transformer to better exploit the mutual benefits between the two tasks. MS-DETR~\cite{ma2025ms} further introduces a motion-semantic disentangled encoder that explicitly separates temporal motion and spatial semantic information, enabling more refined interaction with text queries. More recently, SG-DETR~\cite{gordeev2026saliency} proposes saliency-guided attention for cross-modal interaction, where local saliency scores are used to guide cross-modal attention.

Despite their strong performance, most existing MR\&HD methods primarily focus on visual-language alignment. Audio, when available, is often omitted, treated as a secondary modality, or incorporated through simple feature concatenation. As a result, the effectiveness of audio for highlight detection remains relatively underexplored, especially in sports videos where audio cues are often highly informative.

\textbf{Audio-Visual Highlight Detection.}
Beyond joint MR\&HD, several works have studied highlight detection more directly. JAV~\cite{badamdorj2021joint} is one of the earlier methods to investigate the interaction between visual and audio modalities for highlight detection. It introduces unimodal self-attention and bimodal attention mechanisms to jointly learn from visual and audio information. UMT~\cite{liu2022umt} extends this line of work by considering the reliability of different modalities. Since visual, audio, and textual modalities may be missing or noisy in real-world scenarios, UMT proposes a unified multimodal Transformer to handle different modality combinations and reliability conditions.

Several recent works further investigate highlight detection in more practical or domain-specific settings. Della et al.~\cite{della2025automated} specifically target sports video highlight detection, where audio is first converted into mel-spectrogram representations and then processed using 2D convolutional layers. TTA~\cite{islam2026test} addresses the distribution shift between training and inference videos by introducing test-time adaptation, allowing the model to adapt to the specific characteristics of each video during inference.

In addition to supervised approaches, unsupervised highlight detection has also been explored~\cite{badamdorj2022contrastive,islam2025unsupervised,li2024unsupervised}. These methods suggest that highlight clips often share common feature patterns that can be identified even without explicit supervision. More recently, several works have attempted to address video temporal grounding and highlight-related tasks using large language models or large video-language models in zero-shot or weakly supervised settings~\cite{guo2025vtg,ren2024timechat,guo2024trace,lee2026svhighlights}.

Overall, existing methods have made substantial progress in visual-language moment retrieval and multimodal highlight detection. However, the role of audio itself remains insufficiently examined. In particular, it is still unclear how far audio-only models can go for sports highlight detection, and whether audio should be treated as a primary signal rather than merely an auxiliary modality. In this work, we revisit this question by systematically evaluating audio-only temporal models, comparing them with visual and audio-visual baselines, and analyzing the contribution of different audio sources such as vocal/commentary and background sounds.

\textbf{Dataset.}
Existing video highlight detection benchmarks primarily focus on general-domain content. Representative datasets include YouTube Highlights \cite{sun2014ranking}, TVSum \cite{song2015tvsum}, and QV-Highlights \cite{lei2021detecting}, which jointly supports moment retrieval and highlight detection. These benchmarks are largely composed of diverse YouTube videos covering everyday activities and a broad range of topics, rather than being specifically designed for sports highlight detection.

In contrast, sports video understanding has attracted increasing attention in recent years, leading to the development of numerous sports-specific datasets. Among them, SoccerNet \cite{giancola2018soccernet} is one of the most widely used benchmarks and has subsequently been extended to support a variety of tasks, including action spotting, replay grounding, video summarization, and other forms of soccer video analysis \cite{deliege2021soccernet,cioppa2022scaling,gao2020automatic,rao2025towards,sarkhoosh2024soccersum}.

SoccerHigh \cite{diaz2025soccerhigh} is a sports highlight detection dataset constructed from SoccerNet and contains 237 soccer matches. Its highlight annotations are derived from official highlight compilations published on the official YouTube channels of the corresponding leagues. However, the dataset is limited to a single sport, which restricts its ability to support the study of highlight characteristics across different sports.

More recently, SV-Highlights \cite{lee2026svhighlights} was introduced as a large-scale multi-sport highlight detection benchmark. It contains 320 videos across eight sports: American football, baseball, basketball, ice hockey, racing, rugby, soccer, and volleyball, with 40 videos per sport. Each video has an average duration of approximately two hours, providing a challenging benchmark for highlight detection in long-form and diverse sports videos.

\begin{figure*}
    \centering
    \includegraphics[width=0.8\linewidth]{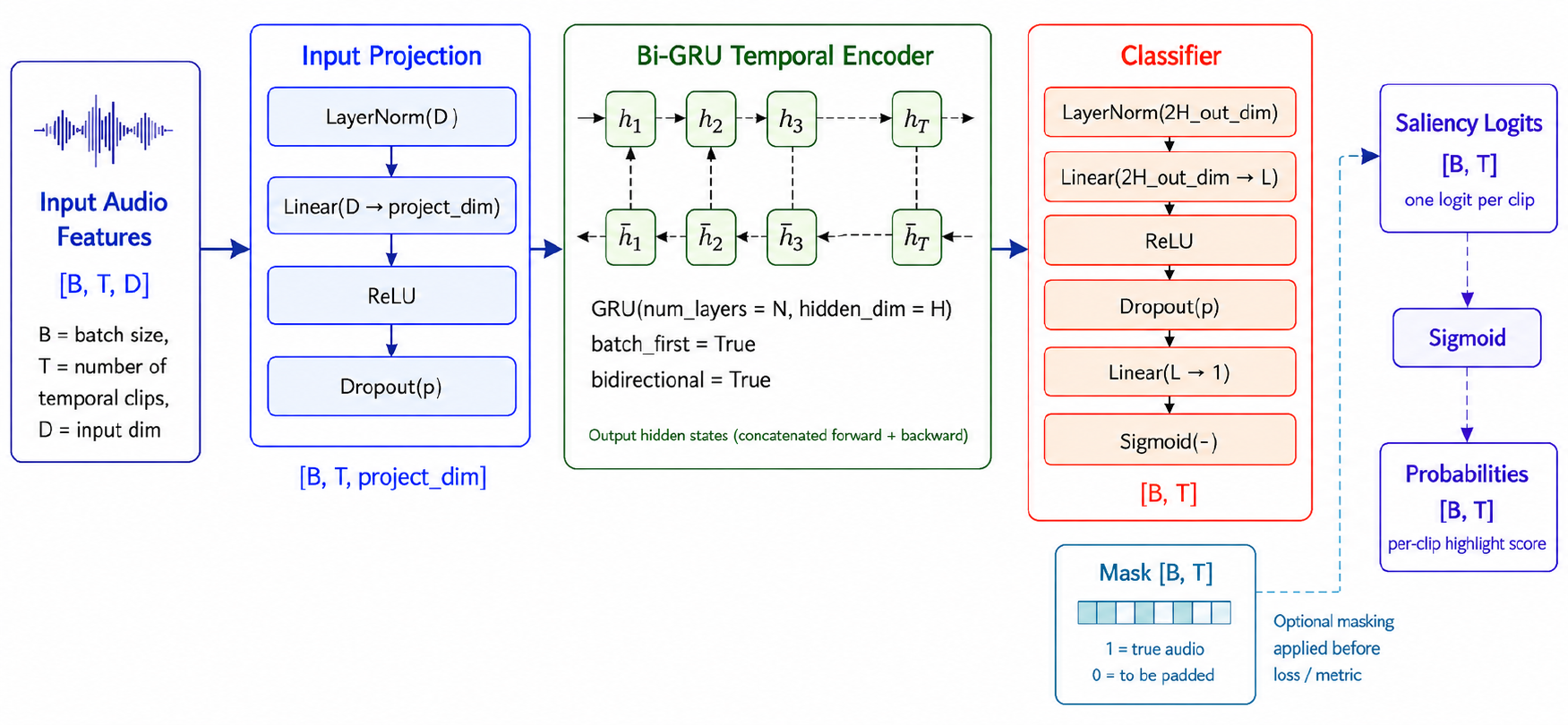}
    \caption{Architecture overview of the Audio GRU baseline.}
    \label{fig:gru_overview}
\end{figure*}

\section{Empirical Study Design}
\label{sec:method}
Our empirical study is organized around the following research questions:
\begin{enumerate}
    \item \textbf{RQ1:} How effective is audio alone for sports highlight detection?
    \item \textbf{RQ2:} How important is temporal modelling for exploiting highlight-relevant audio cues?
    \item \textbf{RQ3:} Which components of sports audio, including vocal and background signals, contribute to highlight detection?
    \item \textbf{RQ4:} Does audio provide complementary information beyond visual features?
\end{enumerate}

To answer these questions, we construct controlled visual-only, audio-only, and audio-visual baselines, compare clip-independent and temporal models, and conduct source-separated audio analysis.

\subsection{Problem Formulation}

Given an untrimmed sports video, we divide it into a sequence of $T$ temporal clips,
\begin{equation}
    \mathcal{V}=\{v_t\}_{t=1}^{T}.
\end{equation}
Each clip is associated with a binary highlight label $y_t \in \{0,1\}$, where $y_t=1$ indicates that the clip belongs to a highlight segment.

For each clip, we extract pretrained visual and audio representations. The visual and audio feature sequences are denoted as
\begin{equation}
    \mathbf{X}^{v}=\{\mathbf{x}^{v}_t\}_{t=1}^{T},
    \qquad
    \mathbf{X}^{a}=\{\mathbf{x}^{a}_t\}_{t=1}^{T}.
\end{equation}
Depending on the experimental setting, the model receives visual features, audio features, or both, and predicts a clip-level saliency score
\begin{equation}
    \hat{s}_t=f(\mathbf{x}_t), \qquad \hat{s}_t \in [0,1].
\end{equation}

\subsection{Evaluating Audio as a Standalone Modality}

To answer \textbf{RQ1}, we evaluate whether audio alone contains sufficient information for sports highlight detection. Each video is represented only by its pretrained audio feature sequence
\begin{equation}
    \mathbf{X}^{a}=\{\mathbf{x}^{a}_{t}\}_{t=1}^{T},
\end{equation}
extracted using the Audio Spectrogram Transformer (AST)~\cite{gong2021ast}. No visual information is provided to the model.

We consider two controlled variants. The first is a clip-independent MLP:
\begin{equation}
    \hat{s}_{t}=\sigma(\mathrm{MLP}(\mathbf{x}^{a}_{t})),
\end{equation}
which measures how much highlight-relevant information is contained within an individual audio clip.

The second introduces temporal modelling as shown in Figure \ref{fig:gru_overview}:
\begin{equation}
    \mathbf{h}^{a}_{1:T}
    =
    \mathrm{GRU}\bigl(\phi_a(\mathbf{X}^{a})\bigr),
\end{equation}
followed by
\begin{equation}
    \hat{s}_{t}
    =
    \sigma\bigl(\mathrm{MLP}(\mathbf{h}^{a}_{t})\bigr),
\end{equation}
where $\phi_a(\cdot)$ is a learnable projection layer.

The comparison between MLP and GRU directly addresses \textbf{RQ2}. The MLP treats clips independently, while the GRU can capture temporal acoustic progression, such as commentator build-up, sudden crowd reactions, whistles, impacts, and post-event excitement. A substantial improvement from MLP to GRU would indicate that audio-based highlight detection relies not only on isolated acoustic cues, but also on their temporal evolution.

\subsection{Audio-Visual Complementarity Test}

To answer \textbf{RQ4}, we examine whether audio contributes information beyond visual features. This experiment is not intended to introduce a novel fusion architecture; instead, it serves as a controlled complementarity test.

Given visual and audio feature sequences
\begin{equation}
    \mathbf{X}^{v}=\{\mathbf{x}^{v}_{t}\}_{t=1}^{T},
    \qquad
    \mathbf{X}^{a}=\{\mathbf{x}^{a}_{t}\}_{t=1}^{T},
\end{equation}
we first project both modalities into a shared hidden space:
\begin{equation}
    \mathbf{z}^{v}_{t}=\phi_v(\mathbf{x}^{v}_{t}),
    \qquad
    \mathbf{z}^{a}_{t}=\phi_a(\mathbf{x}^{a}_{t}).
\end{equation}
The projected features are concatenated:
\begin{equation}
    \mathbf{z}^{va}_{t}
    =
    [\mathbf{z}^{v}_{t};\mathbf{z}^{a}_{t}],
\end{equation}
and the fused sequence is processed by a GRU:
\begin{equation}
    \mathbf{h}^{va}_{1:T}
    =
    \mathrm{GRU}(\mathbf{z}^{va}_{1:T}).
\end{equation}
The final saliency score is
\begin{equation}
    \hat{s}_{t}
    =
    \sigma\bigl(\mathrm{MLP}(\mathbf{h}^{va}_{t})\bigr).
\end{equation}

If the fused model outperforms both the visual-only and audio-only models, this provides evidence that audio is not merely a substitute for vision, but contributes complementary highlight-relevant information.

\subsection{Source-Separated Audio Analysis Protocol}

To answer \textbf{RQ3}, we investigate which components of sports audio contribute most strongly to highlight detection. Sports audio is a mixture of commentator speech, crowd reactions, whistles, impacts, background music, and environmental sounds. To separate these components approximately, we use Demucs~\cite{defossez2019music} to decompose each audio track into a vocal stream and a background stream.

The vocal stream primarily contains commentator speech and verbal reactions, while the background stream primarily contains crowd noise, whistles, impacts, and stadium ambience. Because source separation is imperfect, residual leakage between streams may remain; therefore, the results should be interpreted as evidence about dominant audio components rather than perfectly isolated semantic categories.

We evaluate three settings:
\begin{equation}
    \mathbf{X}^{a}_{\mathrm{vocal}},
    \qquad
    \mathbf{X}^{a}_{\mathrm{bg}},
    \qquad
    \mathbf{X}^{a}_{\mathrm{mix}},
\end{equation}
where $\mathbf{X}^{a}_{\mathrm{vocal}}$ and $\mathbf{X}^{a}_{\mathrm{bg}}$ denote features extracted from the separated streams, and $\mathbf{X}^{a}_{\mathrm{mix}}$ denotes features extracted from the original audio mixture.

The same MLP and GRU models are trained under all three settings. This controlled comparison allows us to assess whether highlight-relevant information is concentrated in speech-related cues, non-vocal event sounds, or their combination.

\subsection{Interpretable Audio Cue Analysis}

In addition to model-based evaluation, we analyze interpretable acoustic properties of highlight and non-highlight clips. For each 2-second clip, we compute RMS loudness, peak loudness, and frequency-band energy statistics. RMS loudness measures average signal intensity, peak loudness captures the strongest instantaneous response, and frequency-band energy examines whether highlight-related differences are concentrated in specific spectral regions.

This analysis is not used during model training. Instead, it serves as a diagnostic tool for examining whether highlight clips exhibit measurable acoustic differences from non-highlight clips. At the same time, overlap between the resulting distributions allows us to assess whether simple acoustic thresholding is sufficient or whether richer learned and temporal representations are required.

\begin{table}[t]
\centering
\caption{Modality ablation on SV-Highlights. We compare visual-only, audio-only, and audio-visual fusion baselines.}
\label{tab:modality_ablation}
\renewcommand{\arraystretch}{1} 
\resizebox{1\linewidth}{!}{
\begin{tabular}{lccccc}
\toprule
\textbf{Model} & \textbf{Input} & \textbf{mAP} & \textbf{Hit@1} & \textbf{Hit@K} & \textbf{IoU} \\
\midrule
Vision (MLP) & V & $19.98 \pm 0.65$ & $35.59 \pm 2.11$ & $24.19 \pm 0.98$ & $14.07 \pm 0.61$ \\
Vision (GRU) & V & $37.99 \pm 0.39$ & $79.57 \pm 4.21$ & $39.68 \pm 0.58$ & $25.48 \pm 0.38$ \\
\midrule
Audio (MLP) & A & $33.55 \pm 1.14$ & $76.96 \pm 3.96$ & $36.43 \pm 1.07$ & $22.68 \pm 0.81$ \\
Audio (GRU) & A & $41.79 \pm 0.68$ & $86.90 \pm 3.35$ & $42.37 \pm 0.62$ & $27.49 \pm 0.51$ \\
\midrule
Fusion (GRU) & V+A & $\mathbf{46.92 \pm 0.71}$ & $\mathbf{91.09 \pm 1.91}$ & $\mathbf{46.07 \pm 0.57}$ & $\mathbf{30.67 \pm 0.43}$ \\
\bottomrule
\end{tabular}
}
\end{table}

\begin{table*}[t]
\centering
\caption{Comparison with existing methods on SV-Highlights. We report mean and standard deviation over three runs.}
\label{tab:main_results}
\resizebox{0.8\linewidth}{!}{
\begin{tabular}{lccccc}
\toprule
\textbf{Model} & \textbf{Input} & \textbf{mAP} & \textbf{Hit@1} & \textbf{Hit@K} & \textbf{IoU} \\
\midrule
JAV~\cite{badamdorj2021joint} & V+A & $32.03 \pm 0.49$ & $71.20 \pm 5.47$ & $34.82 \pm 0.39$ & $21.48 \pm 0.27$ \\
TTA~\cite{islam2026test} & V+A & $32.37 \pm 1.32$ & $71.20 \pm 9.45$ & $35.62 \pm 1.23$ & $22.09 \pm 0.91$ \\
UMT~\cite{liu2022umt} & V+A & $21.08 \pm 5.28$ & $49.27 \pm 15.74$ & $24.54 \pm 5.90$ & $14.35 \pm 3.80$ \\
QD-DETR~\cite{moon2023query} & V+A & $35.42 \pm 0.58$ & $58.62 \pm 6.07$ & $38.28 \pm 1.01$ & $24.25 \pm 0.77$ \\
MS-DETR~\cite{ma2025ms} & V+A & $36.43 \pm 1.62$ & $64.39 \pm 2.47$ & $39.26 \pm 1.47$ & $25.03 \pm 1.13$ \\
TR-DETR~\cite{sun2024tr} & V+A & $37.77 \pm 0.25$ & $74.36 \pm 3.78$ & $40.02 \pm 0.12$ & $25.63 \pm 0.11$ \\
SG-DETR~\cite{gordeev2026saliency} & V+A & $29.54 \pm 1.15$ & $60.19 \pm 6.08$ & $33.15 \pm 1.01$ & $20.25 \pm 0.70$ \\
UniVTG~\cite{lin2023univtg} & V & $18.87 \pm 0.66$ & $31.38 \pm 6.99$ & $22.87 \pm 0.73$ & $13.15 \pm 0.42$ \\
\midrule
Audio baseline (MLP) & A & $33.55 \pm 1.14$ & $76.96 \pm 3.96$ & $36.43 \pm 1.07$ & $22.68 \pm 0.81$ \\
Audio baseline (GRU) & A & $\mathbf{41.79 \pm 0.68}$ & $86.90 \pm 3.35$ & $42.37 \pm 0.62$ & $27.49 \pm 0.51$ \\
Fusion baseline (GRU) & V+A & $\mathbf{46.92 \pm 0.71}$ & $\mathbf{91.09 \pm 1.91}$ & $\mathbf{46.07 \pm 0.57}$ & $\mathbf{30.67 \pm 0.43}$ \\
\bottomrule
\end{tabular}
}
\end{table*}

\section{Experiments}

This section reports the results of our empirical study on the role of audio in sports highlight detection. We first describe the dataset, evaluation protocol, and implementation details. We then organize the analysis around four research questions we mentioned in section \ref{sec:method}.

\subsection{Dataset and Evaluation Metrics}

We evaluate all models on SV-Highlights~\cite{lee2026svhighlights}, a benchmark for highlight detection in extremely long sports videos. The original dataset contains 320 videos across eight sports: American football, baseball, basketball, ice hockey, racing, rugby, soccer, and volleyball, with 40 videos per sport. Each video has an average duration of approximately two hours.

Due to YouTube availability restrictions, three soccer videos could not be downloaded, and the original videos could not be redistributed by the dataset authors. We therefore conduct our experiments on the remaining 317 videos. Following the dataset protocol, each video is divided into non-overlapping 2-second clips. Each clip is assigned a binary label, where 1 indicates that the clip belongs to a highlight segment and 0 otherwise. The ground-truth annotations are derived from the corresponding official highlight videos.

We follow the SV-Highlights \cite{lee2026svhighlights} evaluation protocol and report mean average precision (mAP), Hit@1, Hit@K, and temporal Intersection-over-Union (IoU). mAP evaluates the overall ranking quality of clip-level saliency predictions. Hit@1 indicates whether the highest-ranked predicted clip corresponds to a ground-truth highlight. For Hit@K, $K$ is defined independently for each video as the number of ground-truth highlight clips in that video. Specifically,
\begin{equation}
\mathrm{Hit@K}
=
\frac{
\left|
\mathcal{P}_{K} \cap \mathcal{G}
\right|
}{
\left|
\mathcal{G}
\right|
},
\qquad
K = \left|\mathcal{G}\right|,
\end{equation}
where $\mathcal{P}_{K}$ denotes the set of the top-$K$ highest-scoring predicted clips and $\mathcal{G}$ denotes the set of ground-truth highlight clips. Hit@K therefore measures the proportion of ground-truth highlight clips recovered when the number of selected predictions is matched to the number of ground-truth highlights. IoU measures the temporal overlap between predicted and ground-truth highlight regions.

Although SoccerHigh \cite{diaz2025soccerhigh} is closely related to our task, we were unable to include it in our experiments because the required untrimmed SoccerNet videos were not publicly accessible. The currently downloadable SoccerNet videos are temporally trimmed, whereas SoccerHigh provides highlight annotations and pre-extracted visual features based on the original untrimmed matches. Since our method additionally requires audio features extracted from the raw videos, the audio obtained from the available trimmed videos cannot be temporally aligned with the SoccerHigh annotations. This mismatch prevents a valid and reproducible evaluation on the dataset.

\subsection{Implementation Details}

SV-Highlights does not provide an official train/validation/test split. Since our study requires supervised training, we construct video-level splits. We first allocate 80\% of the videos to training and 20\% to testing, and then use 20\% of the training set for validation. All splits are performed at the video level to prevent clips from the same video appearing in different subsets.

All models are trained for 30 epochs on an NVIDIA A100-SXM4-40GB GPU. We use AdamW~\cite{loshchilov2017decoupled} with an initial learning rate of $1\times10^{-4}$, weight decay of $1\times10^{-4}$, and a cosine learning-rate scheduler. Binary cross-entropy is used as the primary training objective.

Each video is represented using pretrained clip-level features extracted from non-overlapping 2-second segments. We use CLIP~\cite{radford2021learning} for visual representation, AST~\cite{gong2021ast} for audio representation. All feature extractors remain frozen, and only the highlight prediction models are optimized.

During inference, the model outputs clip-level logits, which are converted into saliency probabilities using a sigmoid function. A clip is classified as a highlight when its predicted probability is greater than or equal to 0.5; otherwise, it is classified as a non-highlight clip.

To account for randomness in data splitting and model training, we repeat all experiments using three random seeds (42, 2026, and 3020). For each seed, all methods, including the compared baselines, are retrained and evaluated on the same video-level train/validation/test split using the same modality-specific features, training budget, and metric implementation. For text-query-based methods, we use the fixed query, ``Find the highlight moments in this sports video.'' We report the mean and standard deviation across the three runs.

\begin{table*}[t]
\centering
\caption{Source-separated audio analysis on SV-Highlights. We compare vocal-only, background-only, and original combined audio settings.}
\label{tab:source_separation}
\renewcommand{\arraystretch}{1} 
\resizebox{0.8\linewidth}{!}{
\begin{tabular}{llcccc}
\toprule
\textbf{Model} & \textbf{Audio Input} & \textbf{mAP} & \textbf{Hit@1} & \textbf{Hit@K} & \textbf{IoU} \\
\midrule
Audio baseline (MLP) & Vocal Only & $29.83 \pm 0.83$ & $73.30 \pm 0.24$ & $33.53 \pm 0.68$ & $20.44 \pm 0.49$ \\
Audio baseline (GRU) & Vocal Only & $37.39 \pm 0.95$ & $83.78 \pm 3.54$ & $38.99 \pm 0.34$ & $24.67 \pm 0.28$ \\
\midrule
Audio baseline (MLP) & Background Only & $26.67 \pm 1.32$ & $61.27 \pm 7.29$ & $29.95 \pm 1.36$ & $18.01 \pm 0.95$ \\
Audio baseline (GRU) & Background Only & $33.52 \pm 0.17$ & $68.03 \pm 8.81$ & $35.68 \pm 0.41$ & $22.28 \pm 0.32$ \\
\midrule
Audio baseline (MLP) & Combined & $33.55 \pm 1.14$ & $76.96 \pm 3.96$ & $36.43 \pm 1.07$ & $22.68 \pm 0.81$ \\
Audio baseline (GRU) & Combined & $\mathbf{41.79 \pm 0.68}$ & $\mathbf{86.90 \pm 3.35}$ & $\mathbf{42.37 \pm 0.62}$ & $\mathbf{27.49 \pm 0.51}$ \\
\bottomrule
\end{tabular}
}
\end{table*}

\subsection{How Effective Is Audio Alone?}

Table~\ref{tab:main_results} compares the audio-only models with existing highlight detection approaches on SV-Highlight \cite{lee2026svhighlights}. Under our supervised evaluation protocol, the audio-only GRU achieves $41.79$ mAP, $86.90$ Hit@1, $42.37$ Hit@K, and $27.49$ IoU. It outperforms all compared existing methods in mAP, including several methods that use both visual and audio inputs.

This result is notable because the audio-only model has no access to player appearance, object motion, scene context, or other visual information. Its performance indicates that sports audio contains substantial highlight-relevant evidence, including commentator reactions, audience responses, whistles, impacts, and other event-associated sounds. The audio-only MLP also achieves $33.55$ mAP, showing that individual audio clips already contain useful saliency cues even without explicit temporal modelling.

These findings answer \textbf{RQ1}: audio is not merely an auxiliary signal for sports highlight detection. It can serve as a strong standalone modality and, when temporally modelled, can compete with or outperform substantially more complex multimodal systems.

\subsection{How Important Is Temporal Modelling?}

We evaluate the importance of temporal context by comparing clip-independent MLP models with GRU-based temporal models in Table~\ref{tab:modality_ablation}.

For audio, introducing a GRU improves mAP from $33.55$ to $41.79$, Hit@1 from $76.96$ to $86.90$, Hit@K from $36.43$ to $42.37$, and IoU from $22.68$ to $27.49$. The improvement is substantial across all metrics. A similar trend is observed for vision, where the GRU improves mAP from $19.98$ to $37.99$.

These results show that highlight detection depends strongly on temporal progression rather than isolated clip-level evidence. Sports highlights are often preceded and followed by structured acoustic patterns, such as commentator build-up, rising crowd excitement, a sudden impact or whistle, and post-event celebration. The GRU can capture this evolution, whereas the MLP evaluates each clip independently.

The results therefore answer \textbf{RQ2}: temporal modelling is critical for exploiting both audio and visual information, and it is especially important for audio, where highlight cues frequently unfold over multiple adjacent clips.

\subsection{Which Audio Components Contribute Most?}

\subsubsection{Source-Separated Audio}

To investigate which components of sports audio are most informative, we compare vocal-only, background only, and original mixed audio in Table~\ref{tab:source_separation}.

Vocal-only audio consistently outperforms background-only audio. With the GRU model, vocal-only audio reaches $37.39$ mAP, compared with $33.52$ mAP for background-only audio. The same trend is observed with the MLP, where vocal-only and background-only audio achieve $29.83$ and $26.67$ mAP, respectively.

This suggests that commentary and speech-related reactions carry particularly useful information. Commentators often describe decisive events, change their speaking intensity, and react emotionally to important moments. However, the source-separated result should not be interpreted as proving that semantic speech understanding alone explains the performance, since source separation is imperfect and residual cross-stream information may remain.

Background-only audio nevertheless remains informative. The background GRU achieves $33.52$ mAP and $22.28$ IoU, showing that crowd reactions, whistles, ball impacts, and environmental sounds provide useful saliency cues even without commentary.

The original mixed audio performs best, with the GRU reaching $41.79$ mAP. It improves over vocal-only audio by $4.40$ mAP and background-only audio by $8.27$ mAP. This indicates that vocal and non-vocal audio provide complementary information rather than redundant evidence.

\subsubsection{Interpretable Acoustic Cues}

We further analyse the acoustic properties of highlight and
non-highlight clips using four interpretable descriptors: RMS loudness,
peak loudness, low-mid frequency energy, and high-mid frequency energy.

Highlight clips are louder on average than non-highlight clips. The mean
RMS loudness is $-26.925$ dB for highlight clips and $-28.435$ dB for
non-highlight clips, corresponding to a difference of approximately
$1.51$ dB. Similarly, the mean peak loudness is $-9.016$ dB for
highlight clips and $-10.138$ dB for non-highlight clips, giving a
difference of approximately $1.12$ dB.

Highlight clips also exhibit greater spectral energy in both analysed
frequency ranges. The mean low-mid frequency energy is approximately
$0.76$ dB higher for highlight clips, while the mean high-mid frequency
energy is approximately $2.05$ dB higher. These frequency bands may
capture a combination of commentary, crowd reactions, whistles, ball
impacts, and other sharp event-related sounds.

Despite these average differences, the highlight and non-highlight
distributions overlap substantially across all four descriptors. This
indicates that simple acoustic statistics are informative at the
population level but are insufficient for reliable highlight detection
through fixed thresholding. The strong performance of the AST-GRU model
is therefore unlikely to arise from loudness alone. Instead, it likely
benefits from richer pretrained audio representations together with
temporal modelling across neighbouring clips.

Together with the source-separation results, this analysis addresses
\textbf{RQ3}: highlight-relevant information is distributed across both
vocal and background audio and is reflected in a combination of
semantic, affective, and low-level acoustic cues.

\subsection{Does Audio Complement Visual Information?}

To determine whether audio provides information beyond vision, we compare visual-only, audio-only, and audio-visual models in Table~\ref{tab:modality_ablation}.

The audio-only GRU outperforms the visual-only GRU by $3.80$ mAP, achieving $41.79$ compared with $37.99$. More importantly, the simple audio-visual fusion GRU achieves the strongest results across all metrics: $46.92$ mAP, $91.09$ Hit@1, $46.07$ Hit@K, and $30.67$ IoU.

Fusion improves mAP by $5.13$ points over the audio-only GRU and by $8.93$ points over the visual-only GRU. Similar improvements are observed for IoU, where fusion exceeds audio-only and vision-only by $3.18$ and $5.19$ points, respectively.

These results demonstrate that audio and visual features capture complementary aspects of sports highlights. Visual features provide information about player actions, motion, scene context, and game structure, while audio captures commentary, audience response, whistles, impacts, and post-event reactions. The improvement achieved through simple concatenation suggests that existing highlight detection systems may not fully exploit the available audio information.

The results answer \textbf{RQ4}: audio is valuable not only as a standalone modality, but also as a complementary source of evidence that improves visual highlight detection.

\begin{table}[t]
\centering
\caption{Sport-specific temporal prediction behaviour at a decision
threshold of 0.5. GT denotes the ground-truth positive ratio, and PPR
denotes the predicted positive ratio. Over-prediction is the ratio
between PPR and GT, while Region Ratio is the ratio between the numbers
of predicted and ground-truth highlight regions. The overall results
are computed across the complete test set. Best results are shown in
\textbf{bold}, and second-best results are \underline{underlined}.}
\label{tab:temporal_behaviour}
\small
\setlength{\tabcolsep}{7pt}
\resizebox{1\linewidth}{!}{
\begin{tabular}{llcccc}
\toprule
\textbf{Sport} &
\textbf{Model} &
\textbf{GT (\%)} &
\textbf{PPR (\%)} &
\textbf{Over-pred.} &
\textbf{Region Ratio} \\
\midrule

\multirow{4}{*}{NFL}
& MS-DETR    & 10.27 & 28.41 & 2.77 & 3.01 \\
& Vision GRU & 10.27 & 30.26 & 2.95 & 2.34 \\
& Audio GRU  & 10.27 & \textbf{25.62} & \textbf{2.50} & \textbf{1.83} \\
& Fusion GRU & 10.27 & \underline{26.99} & \underline{2.63} & \underline{2.17} \\
\midrule

\multirow{4}{*}{Baseball}
& MS-DETR    & 4.33 & \underline{19.95} & \underline{4.61} & 2.62 \\
& Vision GRU & 4.33 & 26.06 & 6.02 & 2.17 \\
& Audio GRU  & 4.33 & \textbf{17.01} & \textbf{3.93} & \textbf{1.14} \\
& Fusion GRU & 4.33 & 21.72 & 5.02 & \underline{1.84} \\
\midrule

\multirow{4}{*}{Basketball}
& MS-DETR    & 10.04 & 32.93 & 3.28 & 2.38 \\
& Vision GRU & 10.04 & \textbf{27.19} & \textbf{2.71} & 2.22 \\
& Audio GRU  & 10.04 & \underline{27.40} & \underline{2.73} & \textbf{1.50} \\
& Fusion GRU & 10.04 & 32.97 & 3.28 & \underline{1.88} \\
\midrule

\multirow{4}{*}{Ice hockey}
& MS-DETR    & 7.97 & \underline{22.71} & \underline{2.85} & 3.31 \\
& Vision GRU & 7.97 & 23.44 & 2.94 & 2.50 \\
& Audio GRU  & 7.97 & 29.70 & 3.72 & \underline{2.34} \\
& Fusion GRU & 7.97 & \textbf{22.18} & \textbf{2.78} & \textbf{2.20} \\
\midrule

\multirow{4}{*}{Race}
& MS-DETR    & 8.02 & \textbf{16.44} & \textbf{2.05} & 4.47 \\
& Vision GRU & 8.02 & 17.30 & 2.16 & 2.81 \\
& Audio GRU  & 8.02 & \underline{17.06} & \underline{2.13} & \textbf{2.14} \\
& Fusion GRU & 8.02 & 19.92 & 2.48 & \underline{2.79} \\
\midrule

\multirow{4}{*}{Rugby}
& MS-DETR    & 6.11 & 13.29 & 2.18 & 2.84 \\
& Vision GRU & 6.11 & 21.78 & 3.57 & 2.53 \\
& Audio GRU  & 6.11 & \textbf{11.25} & \textbf{1.84} & \textbf{1.34} \\
& Fusion GRU & 6.11 & \underline{11.26} & \underline{1.84} & \underline{1.71} \\
\midrule

\multirow{4}{*}{Soccer}
& MS-DETR    & 6.41 & \textbf{13.92} & \textbf{2.17} & 2.94 \\
& Vision GRU & 6.41 & 30.58 & 4.78 & 3.05 \\
& Audio GRU  & 6.41 & 18.38 & 2.87 & \underline{2.39} \\
& Fusion GRU & 6.41 & \underline{16.79} & \underline{2.62} & \textbf{2.03} \\
\midrule

\multirow{4}{*}{Volleyball}
& MS-DETR    & 8.07 & \underline{31.63} & \underline{3.92} & 4.57 \\
& Vision GRU & 8.07 & 36.32 & 4.50 & 4.05 \\
& Audio GRU  & 8.07 & \textbf{28.02} & \textbf{3.47} & \textbf{3.12} \\
& Fusion GRU & 8.07 & 37.40 & 4.64 & \underline{3.84} \\
\midrule

\multirow{4}{*}{Overall}
& MS-DETR    & 7.53 & \underline{22.41} & \underline{2.98} & 3.10 \\
& Vision GRU & 7.53 & 26.50 & 3.52 & 2.57 \\
& Audio GRU  & 7.53 & \textbf{21.58} & \textbf{2.87} & \textbf{1.84} \\
& Fusion GRU & 7.53 & 23.71 & 3.15 & \underline{2.24} \\

\bottomrule
\end{tabular}}
\end{table}

\section{Discussion \& Limitations}

Our results consistently show that audio carries substantial
highlight-relevant information. The audio-only temporal baseline
outperforms several existing audio-visual methods under our evaluation
setting, while the source-separation experiments show that both vocal
and background signals contribute useful cues. Vocal audio is generally
more informative, reflecting the semantic and affective value of
commentary, whereas crowd reactions, whistles, impacts, and other
non-vocal sounds provide complementary evidence. The strongest results
are obtained from the original combined audio, indicating that these
sources are most effective when used together.

The comparison between clip-independent MLPs and GRU-based models
further demonstrates the importance of temporal modelling. Across both
audio and visual modalities, incorporating temporal context produces
substantial performance gains. This suggests that highlight detection
depends not only on isolated salient clips, but also on the progression
of events over time, including anticipation, decisive actions, audience
or commentator reactions, and post-event responses.

\paragraph{Over-prediction and limited temporal coherence.}
Although the evaluated models achieve strong ranking-based performance,
their thresholded clip-level predictions remain temporally over-active.
Table~\ref{tab:temporal_behaviour} reports the predicted positive ratio
(PPR), over-prediction ratio (OPR), and region-count ratio (RCR) at a
decision threshold of 0.5, both overall and by sport. Let
$N_{\mathrm{pred}}^{+}$, $N_{\mathrm{gt}}^{+}$, and $N$ denote the
number of predicted highlight clips, ground-truth highlight clips, and
total clips, respectively. We define
\begin{equation}
\mathrm{PPR} =
\frac{N_{\mathrm{pred}}^{+}}{N},
\qquad
\mathrm{GTPR} =
\frac{N_{\mathrm{gt}}^{+}}{N},
\qquad
\mathrm{OPR} =
\frac{\mathrm{PPR}}{\mathrm{GTPR}}
=
\frac{N_{\mathrm{pred}}^{+}}{N_{\mathrm{gt}}^{+}}.
\end{equation}
Thus, $\mathrm{OPR}=1$ indicates that the predicted and annotated
highlight durations are equal, whereas $\mathrm{OPR}>1$ indicates
over-prediction. To characterize temporal fragmentation, we convert
both predictions and annotations into binary temporal sequences and
define a highlight region as a maximal contiguous sequence of positive
clips. Let $N_{\mathrm{pred}}^{\mathrm{reg}}$ and
$N_{\mathrm{gt}}^{\mathrm{reg}}$ denote the numbers of predicted and
ground-truth highlight regions, respectively. The region-count ratio is
defined as
\begin{equation}
\mathrm{RCR} =
\frac{N_{\mathrm{pred}}^{\mathrm{reg}}}
     {N_{\mathrm{gt}}^{\mathrm{reg}}}.
\end{equation}
An $\mathrm{RCR}>1$ therefore indicates that the prediction contains
more temporal highlight regions than the annotation, which is
consistent with increased temporal fragmentation.

Across the complete test set, the ground-truth positive ratio is
7.53\%, whereas the models predict 21.58\%--26.50\% of clips as
highlights, corresponding to over-prediction ratios of 2.87--3.52.
This behaviour occurs across all sports but varies considerably in
severity. Baseball exhibits particularly high over-prediction, with
the ratio reaching 6.02 for the visual GRU, partly because only 4.33\%
of its clips are annotated as highlights. Volleyball also produces
consistently high ratios of 3.47--4.64. By contrast, the least
over-active models for rugby, race, and soccer predict approximately
1.84--2.17 times the annotated highlight duration. This variation
suggests that a fixed decision threshold does not transfer uniformly
across sports with different highlight frequencies and temporal
structures.

The predictions are also frequently fragmented. Overall, MS-DETR has
the highest region-count ratio at 3.10, indicating that it predicts
more than three times as many temporal regions as are annotated, while
the audio GRU has the lowest ratio at 1.84. The audio GRU produces the
lowest RCR in seven of the eight sports, including baseball (1.14),
rugby (1.34), and basketball (1.50); the exception is ice hockey,
where the fusion GRU achieves the lowest RCR at 2.20. Nevertheless,
even the least fragmented models generally produce more highlight
regions than are annotated.

Over-prediction and fragmentation capture distinct temporal
behaviours. For example, MS-DETR has a relatively low over-prediction
ratio of 2.05 on race videos but a region-count ratio of 4.47,
indicating that its predicted highlight duration is distributed across
a substantially larger number of temporal regions. Conversely, the
audio GRU generally produces fewer and more temporally coherent
regions, although it still predicts substantially more highlight
content than the ground truth. These behaviours are not captured by
ranking-based metrics such as mAP and Hit@1.

The observed over-activation may have two explanations. Models may
respond to salient but non-highlight events, such as excited
commentary, crowd reactions, whistles, replays, or near-miss actions.
Alternatively, because the annotations represent editorially selected
summaries rather than exhaustive records of salient events, some
apparent false positives may correspond to meaningful but unannotated
moments. Regardless, the results demonstrate that strong clip-level
ranking performance does not necessarily translate into temporally
compact and coherent highlight summaries. Future work could therefore
investigate region-level prediction, continuity-aware objectives,
structured temporal decoding, sport-adaptive calibration, and the
suppression of repeated or weakly supported saliency peaks.

Our study has several additional limitations. The evaluated baselines
are intentionally simple because our aim is to revisit the role of
audio rather than propose a complex architecture. Stronger audio models
could incorporate speech transcription, commentary semantics, explicit
audio event detection, or audio-language models. Similarly, our
concatenation-based fusion could be extended through reliability-aware
fusion, temporal-offset modelling, and adaptive audio-visual weighting,
particularly because modality reliability may vary across sports.
Finally, SV-Highlights has no official train/validation/test split, so
our supervised evaluation relies on constructed video-level splits;
standardized splits would improve reproducibility and comparability.

Overall, the strong audio-only and audio-visual results demonstrate that
audio should be treated as a primary source of information for sports
highlight detection. At the same time, the observed sport-dependent
over-prediction and fragmentation show that stronger region-level
temporal modelling, adaptive multimodal fusion, and sport-aware
calibration remain necessary for producing reliable highlight
summaries.

\section{Conclusion}

This paper presented a controlled empirical study of audio for
sports highlight detection. We showed that a lightweight audio-only
GRU is highly competitive with existing multimodal methods, while
simple audio-visual fusion achieves the strongest overall performance.
Source-separated and acoustic analyses further indicate that
commentary, crowd response, impact sounds, and other acoustic cues
provide complementary forms of highlight-relevant information.

These findings suggest that audio should be treated as a primary
modality in sports highlight detection rather than merely as an
auxiliary signal. Future work should focus on semantic commentary
understanding, temporally coherent region prediction, audio-visual
alignment, and more adaptive multimodal fusion.

\clearpage

\bibliographystyle{ACM-Reference-Format}
\balance
\bibliography{Source/sample-base}

\end{document}